\documentclass[twoside]{article}
\usepackage{fancyhdr}
\usepackage[numbers]{natbib}
\usepackage[textwidth=6in,left=1.25in,right=1.25in,asymmetric]{geometry}
\usepackage{amsmath}
\usepackage{mathrsfs}

\usepackage{mathabx}

\usepackage{amsthm}

\usepackage{amsfonts}
\usepackage{amssymb}
\usepackage{graphicx}
\usepackage{placeins}
\usepackage{algorithm2e}
\usepackage{caption}
\usepackage{array}
\usepackage{multicol}
\usepackage{color}
\usepackage{bbm}
\usepackage{enumerate}
\usepackage{tikz}
\usepackage{mhequ}
\usepackage[font={small,it}]{caption}

\theoremstyle{plain}

\newtheorem{theorem}{Theorem}[section]

\theoremstyle{definition}
\newtheorem{definition}[theorem]{Definition}

\newcommand{\bb}[1]{\mathbb{#1}}

\newcommand{\tx}{\widetilde{x}}
\newcommand{\tz}{\widetilde{z}}
\newcommand{\tZ}{\widetilde{Z}}
\newcommand{\wh}[1]{\widehat{#1}}

\newcommand{\be}{\begin{equs}}
	\newcommand{\ee}{\end{equs}}

\newcommand{\fair}{fair} 

\newcommand{\tX}{\widetilde{X}}

\newcommand{\tF}{\widetilde{F}}

\begin{document}
\allowdisplaybreaks

\pagestyle{fancy}
\fancyhead[RO,LE]{\small\thepage}
\fancyhead[LO]{K. Lum and J. Johndrow}
\fancyhead[RE]{Fair predictive algorithms}

\title{A statistical framework for fair predictive algorithms}
\author{Kristian Lum \\ Human Rights Data Analysis Group \\ \texttt{kl@hrdag.org} 
\and James E. Johndrow \\ Department of Statistics, Stanford University\\ \texttt{johndrow@stanford.edu}
}

\maketitle

\section{Introduction}
Statistical and machine learning algorithms are increasingly used to inform decisions that have large impacts on individuals' lives. Examples include hiring \citep{hoffman2015discretion}, predictive policing \citep{rosenblat2014predicting}, pre-trial risk assessment of recidivism\citep{dieterich2016compas,brennan2009evaluating}, and risk of violence while incarcerated \citep{cunningham2006actuarial}. In many of these cases, the outcome variable to which the predictive models are trained is observed with bias with respect to some legally protected classes. For example, police records do not constitute a representative sample of all crimes \citep{mosher2010mismeasure}.  In particular, black drug users are arrested at a rate that is several times that of white drug users despite the fact that black and white populations are estimated by public health officials to use drugs at roughly the same rate \citep{langan1995racial}. Algorithms trained on such data will produce predictions that are biased against groups that are disproportionately represented in the training data. 

Several approaches have been proposed to correct unfair predictive models. The simplest approach is to exclude the protected variable(s) from the analysis, under the belief that doing so will result in ``race-neutral" predictions \citep{zeng2015interpretable}. Of course, simply excluding a protected variable is insufficient to avoid discriminatory predictions, as any included variables that are correlated with the protected variables still contain information about the protected characteristic. In the case of linear models, this phenomenon is well-known, and is referred to as omitted variable bias \citep{clarke2005phantom}.  Another approach that has been proposed in the computer science literature is to remove information about the protected variables from the set of covariates to be used in predictive models \cite{feldman2015certifying, calders2010three}.  A third alternative is to modify the outcome variable. For example, \cite{kamiran2009classifying} use a naive Bayes classifier to rank each observation and perturb the outcome such that predictions produced by the algorithm are independent of the protected variable. A discussion of several more algorithms for binary protected and outcome variables can be found in \cite{kamishima2013independence}. 

The approach we propose is most similar to \cite{feldman2015certifying}, though we approach the problem from a statistical modeling perspective. We define a procedure consisting of a chain of conditional models.  Within this framework, both protecting and adjusting variables of arbitrary type becomes natural. Whereas previous work has been limited to protecting only binary or categorical variables and adjusting a limited number of covariates, our proposed framework allows for an arbitrary number of variables to be adjusted and for each of these variables and the protected variables to be continuous or discrete. This greatly extends the range of datasets that can be adjusted and thus expands the range of problems to which such adjustments may be applied. Furthermore, because our method uses chained conditional models, practitioners can rely on a vast array of likelihood-based regression methods to implement our approach. For example, using previously proposed methods, if one class in a discrete protected variable had very little data, adjustments would be difficult. In our proposed framework, statistical models for handling data sparsity, such as Bayesian hierarchical models, can easily be deployed to overcome this problem.

\section{Methods} \label{sec:method}
\subsection{Setup}
Suppose we have a response $y$ and predictors $(x,z)$, where $z$ represent protected characteristics. We take $x,z$ to be realizations of $d_x$ and $d_z$ dimensional random vectors $X,Z$ with arbitrary measurement scale. Consider a generic prediction rule or model for $Y$ given by 
$f : x \to \wh y$.

Our goal is not to use any information about $Z$ in predicting $Y$; that is, we want
\begin{definition}[fair  prediction rule]
A prediction is \emph{\fair} with respect to the protected characteristics $Z$ if and only if
$\widehat{Y} \perp Z$. 
\end{definition}
In order to guarantee fair predictions with respect to $Z$ for models fit to $X$,  it is sufficient to define an adjusted $\tx = g(x,z)$ such that  $\tX \perp Z$. Thus, we seek to define a new random variable $\tX$ that is independent of $Z$, while still preserving as much ``information'' in $X$ as possible. The next section is concerned with defining $\tX$.

\subsection{Transformations to independence}
We propose univariate transformations of the form $g(x,z) =\tF^{-1}(F_{x \mid z}(x))$, where $F_{x \mid z}$ is the conditional cumulative distribution function of $X$ given $Z$.  This is a  generalization of the transformation that is theoretially motivated in \cite{feldman2015certifying} (they use a specific form for $\tilde{F}$). Under this tranformation, as long as an adequate conditional model for $X$ can be defined, $F_{x \mid z}(X \mid Z) \sim \text{Uniform}(0,1)$ and $F_{x \mid z}(X \mid Z) \perp X$.\footnote{ If $X$ is discrete, we can still achieve this using a random map constructed from uniforms restricted to intervals depending on $z$.} Defining conditional models, $F_{x \mid z}$, is, in essence, a likelihood-based regression modeling problem and a primary objective of the field of statistics. For example, one could define the conditional distribution of $X$ given $Z$ using a simple Gaussian linear model, i.e. $X = \alpha + \beta Z  + \epsilon$ where $\epsilon \sim N(0, \sigma^2)$. Then, $F_{x \mid z}(x\mid z) = \Phi \left ( \frac{x - \alpha- \beta z}{\sigma} \right )$, where $\Phi$ is the standard Normal distribution function. Of course, in many cases, Gaussian residuals may not be appropriate-- whether simply because the residuals are non-Gaussian or because $X$ is not continuous. In these cases, non-parametric regression methods (e.g. kernel density regression) or generalized linear models (e.g. Poisson regression for count variables) can be employed to similar effect.  By applying this or a similar transformation independently to each of the covariates $X$ in the model, one can achieve pairwise independence with $Z$. This is insufficient to {\it guarantee} fair predictions, though depending on the prediction models used as well as the dependence structure in the data, univariate transformations may be adequate. 


%
%
%
%
%
An analogous multivariate transformation function $g$ can be constructed as
\be \label{eq:chainmap}
g(x,z) = (g_1(x_1,\tz^{(1)}),\ldots,g_{d_x}(x_{d_x},\tz^{(d_x)})) 
\ee
where $\tz^{(j)} := \{z,\tx_{1:j-1}\}$ for $j>1$ and $\tz^{(1)} := \{z\}$. The ordering $x_1,\ldots,x_{d_x}$ of the $x$ variables is arbitrary, though some orderings may be practically convenient for a given application.

\noindent Using basic rules of conditional probability, $p(\tx \mid z)$ can be decomposed as
\be 
\prod_j p(\tx_j \mid \tx_{1:(j-1)},z) = 
\prod_j p(g_j(x_j,\tz^{(j)}) \mid \tz^{(j)}).
\ee
Because $g_j(X_j, \tZ^{(j)}) \perp \tZ^{(j)}$,  each element of the product can be replaced with $p(g_j(x_j, \tz^{(j)})) = p(\tx_j)$ and the joint distribution reduces to  $p(\tx \mid z) = \prod_j p(\tx_j)$. Thus,  $\tX$, $Z$ are mutually independent.  

Although $\tX$ is independent of $Z$, it is important to note that $\tX$ is not independent of $X$. The map in \eqref{eq:chainmap} preserves information in $x$ by maintaining conditional ranks -- if $F_{x_j \mid \tz^{(j)}}(x_{j}) > F_{x_j \mid \tz^{(j)}}(x_j')$, then $\tx_j > \tx_j'$. 

At this point we have defined all of the elements of a viable procedure, with the exception of specifying $\tF$. The choice of $\tF$ does not affect the ranks of $\tx$ at all, so any prediction rule that depends only on the ranks of the predictors will be invariant to $\tF$. This includes regression-tree procedures, such as random forests. Moreover, it is typical in applied statistics and regression modeling to transform predictors prior to model fitting for computational reasons or to obtain better predictive accuracy, which would neutralize any choice we make for $\tF$. On balance, we suggest taking $\tF$ to be the marginal distribution $F_x$. This ensures that researchers using the transformed data still have access to the original marginal distribution of the data, which may be of significant value in its own right.


\section{Application: Removing racial bias in the recidivism risk assessment} \label{sec:application}
Propublica recently compiled an extensive dataset from the criminal justice system in Broward County, Florida to investigate whether risk assessment tools were disproportionately recommending non-release for Black defendants (\cite{angwin2016machine}). For each defendent in the time period, they collected several measures of criminal history-- the number of juvenile misdameanor, felony, and other offenses (denoted by juv\_misd\_count, juv\_fel\_count, juv\_other\_count), number of adult prior offenses (prior\_count)--, the defendant's race, sex, and age at the time of the alleged crime, and an indicator of whether the the defendant was re-arrested within two years of their release. Using only this data, we attempt to build a {\it fair} predictive model in which the outcome of interest is re-arrest within two years. 

We construct regression models for each $x_j$ in the recidivism data, conditional on the protected variable ($z$ = race) and each of the previously transformed variables ($\tx_{1:(j-1)}$). Of the six $x_j$, one (sex) is binary, one ($\log(\text{age})$ -- henceforth simply ``age'') is continuous, and the other four, which relate to prior criminal record, are counts. 
For count variables, we use zero-inflated Poisson or negative binomial regression models to estimate $F_{x \mid \tz}$. Binary $x_j$ are modeled using logistic regression, and continuous $x_j$ using linear regression. For linear regression, we estimate a linear mean function in the usual way, but bootstrap the empirical distribution of the residuals to obtain an estimate of the error distribution. Thus, our model assumes only that the residuals are independent and that the conditional expectation $\bb E[x_j \mid \tz^{(j)}]$ is a linear function. When the likelihood is discrete, as in the case of binary or count data, we sample $q(x) \sim \text{Uniform}(\wh F_{x_j \mid \tz^{(j)}}(x_{-}),  \wh F_{x_j \mid \tz^{(j)}}(x))$, where for an observed value $x$ of $x_j$, $x_{-} = \max_{x'} \{x' : \bb P[x_j = x'] > 0, x' < x\}$, then compute $\wh F_{x_j}^{\leftarrow}(q(x))$ using the empirical quantile function. This random map also achieves fairness. 

The resulting $\tx_j$ is stochastic. While any $(y,\tx)$ generated in this way is fair for race, individual predictions depend on the sampled values $q(x)$ for all of the discrete variables, and interval estimates of parameters will understate uncertainty resulting from the stochastic nature of the maps $g_j$. Consequently, in generating predictive values for individual subjects or estimating uncertainty in model parameters, we use an average over $M$ fair datasets $(y,\tx_j)$. This approach is also used in multiple imputation and privacy settings. To assess fit of the conditional models, we compute the Kolmogorov-Smirnov statistics and associated $p$-values under the null distribution that $F_{x \mid z}(x \mid z)$ (or $q(x)$, in the discrete case) is uniform on the unit interval. In all but one case (priors\_juv\_count), we fail to reject the null hypothesis that $F_{x \mid z}(x \mid z)$ are uniform, suggesting that the conditional models are close to the true conditional distributions of $X \mid Z$.

\begin{figure*}[h!]
\centering
\includegraphics[width=.75\textwidth]{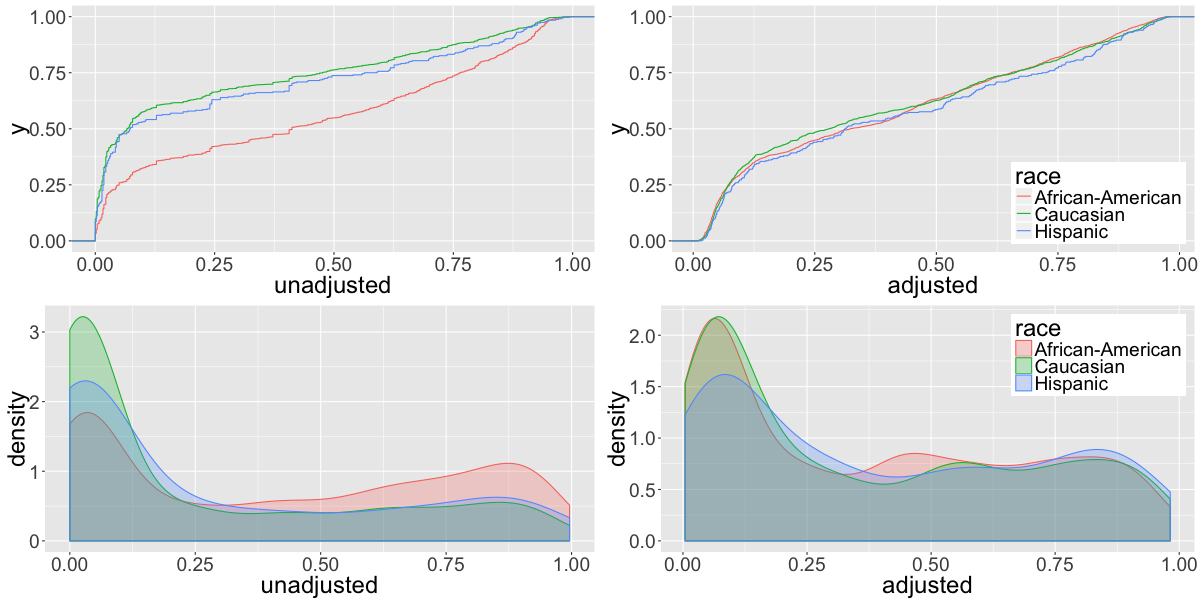}
\caption{density and cdf of predictions made using random forest by race using adjusted and unadjusted data} \label{fig:cdf-rf}
\end{figure*}

%

\begin{figure}[h]
\centering
\includegraphics[width=0.4\textwidth]{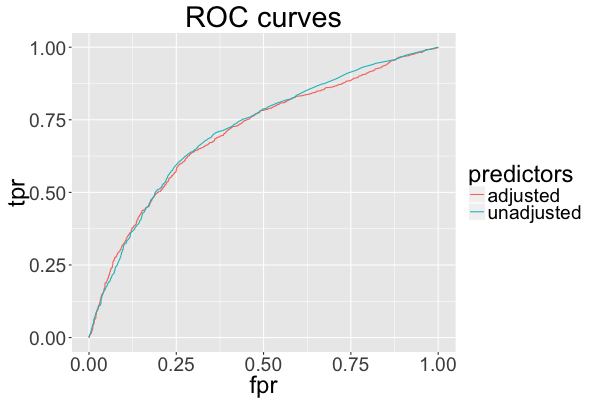}
\caption{roc for predictions made with random forest using adjusted and unadjusted data} \label{fig:roc-rf}
\end{figure}

We then use each of the $M$ transformed datasets to predict re-arrest within two years using random forest (RF). For comparison, we trained RF to the non-transformed data, omitting race as a covariate. Figure \ref{fig:cdf-rf} shows the empirical density and cdf of the re-arrest probability for RF trained on data adjusted using our procedure (adj) and on unadjusted data (unadj). It is clear from the left panels of Figure \ref{fig:cdf-rf} that when trained on unadjusted data, large differences by race exist in the predictive distribution, with the distribution for black individuals having substantially more mass at probabilities of re-arrest greater than about $0.5$. In other words, when trained on unadjusted data omitting race, the predictions of RF are biased against black individuals. Predictions made by training RF on data adjusted using our procedure eliminate almost all racial disparities, as evidenced by the nearly identical distributions by race in the two panels on the right. In applying out procedure, some relevant information is lost, as race is correlated with which defendants are re-arrested. Thus, it is expected that the predictive accuracy of a model fit to the adjusted data will be lower than the model trained on unadjusted data. Figure \ref{fig:roc-rf} shows the ROC curves for both the predictions from the adjusted and unadjusted data. We find that these are not substantially different. For the unadjusted data, the area under the curve (AUC) was 0.71, while for the adjusted data, it was 0.72. We note that this AUC is on par with and, in fact, slightly better than that reported by the makers of a real deployed risk assessment algorithm for this dataset (0.70). \cite{dieterich2016compas}

\section{Discussion}
We have presented a statistical framework for removing information about a protected variable from a datset. Using a statistical modeling approach, we have demonstrated how to apply such adjustments in a general framework. When applied to a dataset of recidivism, our approach successfully created predictions of recidivism that are independent of the protected variable, with minimal loss in predictive accuracy, which was comparable to commercially available algorithms. 



%
\bibliographystyle{plainnat}
\bibliography{rnl_short}

\end{document}